\documentclass{article}
\usepackage{ijcai25}

\usepackage{times}
\usepackage{soul}
\usepackage{url}
\usepackage[hidelinks]{hyperref}
\usepackage[utf8]{inputenc}
\usepackage[small]{caption}
\usepackage{graphicx}
\usepackage{amsmath}
\usepackage{amsthm}
\usepackage{amssymb}
\usepackage{booktabs}
\usepackage{algorithm}
\usepackage{algorithmic}
\usepackage[switch]{lineno}
\usepackage{xcolor}
\usepackage{float}
\usepackage{multirow}
\usepackage{makecell}
\usepackage{tikz}
\usepackage[edges]{forest}
\usepackage{pifont}
\usepackage{mydef}
\usepackage{marvosym}
\newcommand{\cmark}{\ding{51}}
\newcommand{\xmark}{\ding{55}}

\newcommand{\sw}[1]{\textcolor{black}{#1}}

\title{Grounding Creativity in Physics: A Brief Survey of Physical Priors in AIGC}

\author{
Siwei Meng$^1$\and
Yawei Luo$^{2}$\textsuperscript{\Letter} \and
Ping Liu$^1$\textsuperscript{\Letter} \\
\affiliations
$^1$Department of Computer Science, University of Nevada, Reno\\
$^2$School of Software Technology, Zhejiang University
\emails
siweim@unr.edu,
yaweiluo@zju.edu.cn,
pino.pingliu@gmail.com
}

\begin{document}

\maketitle
\footnotetext[1]{\textsuperscript{\Letter}~denotes the co-corresponding author.}

\begin{abstract}
   Recent advancements in AI-generated content have significantly improved the realism of 3D and 4D generation. 
However, most existing methods prioritize appearance consistency while neglecting underlying physical principles, leading to artifacts such as unrealistic deformations, unstable dynamics, and implausible objects interactions. 
Incorporating physics priors into generative models has become a crucial research direction to enhance structural integrity and motion realism.
This survey provides a review of physics-aware generative methods, systematically analyzing how physical constraints are integrated into 3D and 4D generation. 
First, we examine recent works in incorporating physical priors into static and dynamic 3D generation, categorizing methods based on representation types, including vision-based, NeRF-based, and Gaussian Splatting-based approaches. 
Second, we explore emerging techniques in 4D generation, focusing on methods that model temporal dynamics with physical simulations. 
Finally, we conduct a comparative analysis of major methods, highlighting their strengths, limitations, and suitability for different materials and motion dynamics.
By presenting an in-depth analysis of physics-grounded AIGC, this survey aims to bridge the gap between generative models and physical realism, providing insights that inspire future research in physically consistent content generation. 

\end{abstract}

\begin{table*}[htbp]
    
    \setlength{\tabcolsep}{12pt} 
    \renewcommand{\arraystretch}{1.15} 
    \centering
    \scalebox{0.84}{
    \begin{tabular}{c|c|c|c|c}
    \toprule
         \textbf{Works}& \textbf{Focused Tasks} & \textbf{Physics Priors}& \textbf{4D Generation}&\textbf{Released Date} \\
    \midrule
         \cite{cao2023comprehensive_arxiv2023}      & Generative Models in AIGC & \xmark & \xmark & Mar 2023     \\
         \cite{li2024advances_arxiv2024}        & 3D Generation             & \xmark & \xmark & Jan 2024     \\
         \cite{liu2024comprehensive_arxiv2024}     & 3D Generation             & \xmark & \xmark & Feb 2024     \\
         \cite{banerjee2024physics_acmcomput2024} & Physics-informed Computer Vision Models       & \cmark & \xmark & Oct 2024     \\
         \cite{liu2025generative_arxiv2025}      & Physical Simulation with Generative Models  & \cmark & \xmark & Jan 2025     \\
        \textbf{Ours} & \textbf{Physics-aware AIGC}  & \textbf{\cmark} & \textbf{\cmark}  &\textbf{Feb 2025} \\
    \bottomrule
    \end{tabular}}
    \caption{Comparison between this work and previous surveys.}
    \label{tab:survey_comparison}
\end{table*}

\section{Introduction}
Recent advancements in AI-generated content (AIGC) have significantly enhanced 3D and 4D content generation, with applications spanning gaming, simulation, animation, and robotics. 
Traditional AI-driven 3D generation methods primarily focus on improving geometric fidelity and rendering efficiency, leveraging representations such as Neural Radiance Fields (NeRF) \cite{mildenhall2021nerf_eccv2020} and 3D Gaussian Splatting (GS) \cite{kerbl20233d_tog2023}.
More recent models, such as DreamFusion~\cite{poole2022dreamfusion_iclr2023}, integrate Diffusion Models (DMs) to improve synthesis realism. 
However, these models primarily optimize for visual quality, often neglecting physical plausibility, leading to artifacts such as implausible deformations, unstable motion, and inconsistent object interactions.

The integration of physics priors into generative models is an emerging yet underdeveloped direction in 3D/4D generation. Most generative models are trained on datasets lacking explicit physical constraints, failing to capture material properties, object dynamics, and force interactions. 
Consequently, generated content often deviates from real-world physical laws, limiting its applicability in simulation-driven applications. 
To address this gap, recent research has explored differentiable physics-based models such as Material Point Method (MPM) \cite{jiang2016material_siggraph2016,hu2018moving_tog2018}, Finite Element Method (FEM) \cite{zienkiewicz2000finite_2000}, and differentiable physics engines into the generative pipeline. 
These approaches enable physics-informed content generation, ensuring structural integrity, dynamic realism, and physically consistent interactions.
However, a systematic review of these physics-based advances in generative models remains absent. 

Existing surveys on 3D and 4D generation primarily focus on three aspects (\autoref{tab:survey_comparison}): (1) scene representations and rendering \cite{li2024advances_arxiv2024}, which discuss different 3D representations and rendering optimizations; (2) generative models \cite{cao2023comprehensive_arxiv2023,liu2024comprehensive_arxiv2024}, which analyze synthesis techniques from text and images; and (3) applications and scalability \cite{li2024advances_arxiv2024}, covering gaming, animation, and robotics. 
However, none of these works systematically explore the role of physics priors in generative models. 
A recent survey \cite{liu2025generative_arxiv2025} provides the closest discussion to our work, categorizing physics-aware generation into explicit physics simulation-based (PAG-E) and implicit physics-informed (PAG-I) methods.
However, its focus remains on physics-aware video and 3D content generation, with limited discussion on 4D generation and dynamic scene modeling. 

To bridge this gap, this survey provides an overview of physics-grounded generative models, categorizing recent advances into static 3D generation, dynamic 3D generation, and 4D generation. 
Specifically, we further organize existing methods based on representation types and generation paradigms, covering vision-based, NeRF-based, and GS-based dynamic 3D generation approaches (\autoref{fig:taxonomy}).~By synthesizing insights from physics-based simulation and generative models, we aim to establish a systematic perspective on integrating physics priors with AI-driven content generation, offering new directions for research in this evolving field.
A curated list of all related papers mentioned in this work can be found at:~\href{https://github.com/mengsiwei/Awesome-Physical-AIGC-lists}{https://github.com/mengsiwei/Awesome-Physical-AIGC-lists}.

\tikzstyle{leaf}=[draw=hiddendraw,
    rounded corners,minimum height=2.4em,
    fill=myblue!30,text opacity=1, align=center,
    fill opacity=.5,  text=black,align=center,font=\scriptsize,
    inner xsep=6pt,
    inner ysep=3pt,
    ]
\tikzstyle{middle}=[draw=hiddendraw,
    rounded corners,minimum height=1.2em,
    fill=output-white!40,text opacity=1, align=center,
    fill opacity=.5,  text=black,align=center,font=\scriptsize,
    inner xsep=3pt,
    inner ysep=1pt,
    ]
\begin{figure*}[ht]
\centering
\begin{forest}
  for tree={
  forked edges,
  grow=east,
  reversed=true,
  anchor=base west,
  parent anchor=east,
  child anchor=west,
  base=middle,
  font=\scriptsize,
  rectangle,
  line width=0.7pt,
  draw=output-black,
  rounded corners,align=left,
  minimum width=2em,
    s sep=5pt,
    inner xsep=3pt,
    inner ysep=1pt,
  },
  where level=1{text width=4.5em}{},
  where level=2{text width=6em,font=\scriptsize}{},
  where level=3{font=\scriptsize}{},
  where level=4{font=\scriptsize}{},
  where level=5{font=\scriptsize}{},
  [Physical Priors Grounded Models, middle,rotate=90,anchor=north, text width=11em, edge=output-black, draw=mytitle-edge, minimum height=1.3em
    [Static 3D Generation, middle, edge=output-black,text width=8.2em, draw=myyellow-edge
        [Phy3DGen \cite{xu2024precise-physics_arxiv2024}{,} PhyCAGE \cite{yan2024phycage_arxiv2024}{,} Atlas3D \cite{chen2024atlas3d_arxiv2024}{,} PhiP-G \cite{Li2025PhiP-G_arxiv2025}{,} \\ PhysComp3D \cite{guo2024physically_arxiv2024}{,} LAYOUTDREAMER \cite{zhou2025layoutdreamer_arxiv2025}, leaf, text width=36.68em, edge=output-black, fill=myyellow!60, draw=myyellow-edge]
    ]
    [Dynamic 3D Generation, middle, edge=output-black, text width=8.2em, draw=mygreen-edge
        [Vision-based, middle, text width=4.2em, edge=output-black, draw=mygreen-edge
            [Physics 101 \cite{wu2016physics101_bmvc2016}{,} Generative Image Dynamic \cite{li2024generative_cvpr2024}{,} \\ DANO \cite{le2023differentiable_RAL2023}{,} PhysGen \cite{liu2024physgen_arxiv2024}{,}   PhyT2V \cite{xue2024phyt2v_cvpr2025}{,} \\MOTIONCRAFT \cite{savant2024motioncraft_arxiv2024}, leaf, text width=30.7em, edge=output-black, fill=mygreen!90, draw=mygreen-edge]
        ]
        [NeRF-based, middle, text width=4.2em, edge=output-black, draw=mygreen-edge
            [ParticleNeRF  \cite{abou2024particlenerf_wacv2024}{,} PAC-NeRF \cite{lipac-nerf_iclr2023}{,} LPO \cite{kaneko2024improving_cvpr2024}{,} \\ PIE-NeRF \cite{feng2024pie-nerf_cvpr2024}, leaf, text width=30.7em, edge=output-black, fill=mygreen!90, draw=mygreen-edge]
        ]
        [GS-based, middle, text width=4.2em, edge=output-black, draw=mygreen-edge
            [Spring-Gaus  \cite{zhong2024reconstruction_arxiv2024}{,} PhysDreamer \cite{zhang2025physdreamer_eccv2024}{,} GIC \cite{cai2024gic_NeurIPS2024}{,}  \\PhysGaussian \cite{xie2024physgaussian_cvpr2024}{,}   Physics3D \cite{liu2024physics3d_arxiv2024}{,} NeuMA \cite{cao2024neural_arxiv2024}{,} \\ GSP \cite{feng2024splashing_arxiv2024}{,} DreamPhysics \cite{huang2024dreamphysics_arxiv2024}{,}  PhysMotion \cite{tan2024physmotion_arxiv2024}{,} \\ Gaussian Splashing \cite{mualem2024gaussian_arxiv2024}{,} OMNIPHYSGS \cite{lin2025omniphysgs_iclr2025}, leaf, text width=30.7em, edge=output-black, fill=mygreen!90, draw=mygreen-edge]
        ]
    ]
    [4D Generation, middle, edge=output-black, text width=8.2em, draw=mycyan-edge
        [Phy124 \cite{lin2024phy124_arxiv2024}{,} GASP \cite{borycki2024gasp_arxiv2024}{,} TRANS4D \cite{zeng2024trans4d_arxiv2024}{,} Phys4DGen \cite{lin2024phys4dgen_arxiv2024}{,}  \\Unleashing \cite{liu2024unleashing_arxiv2024}{,} NS-4Dynamics \cite{Wang2025Compositional_iclr2025}, leaf, text width=36.68em, edge=output-black, fill=mycyan!40, draw=mycyan-edge]
    ]
  ]
\end{forest}
\caption{A taxonomy of generative models grounded with physical priors.}
\label{fig:taxonomy}
\end{figure*}
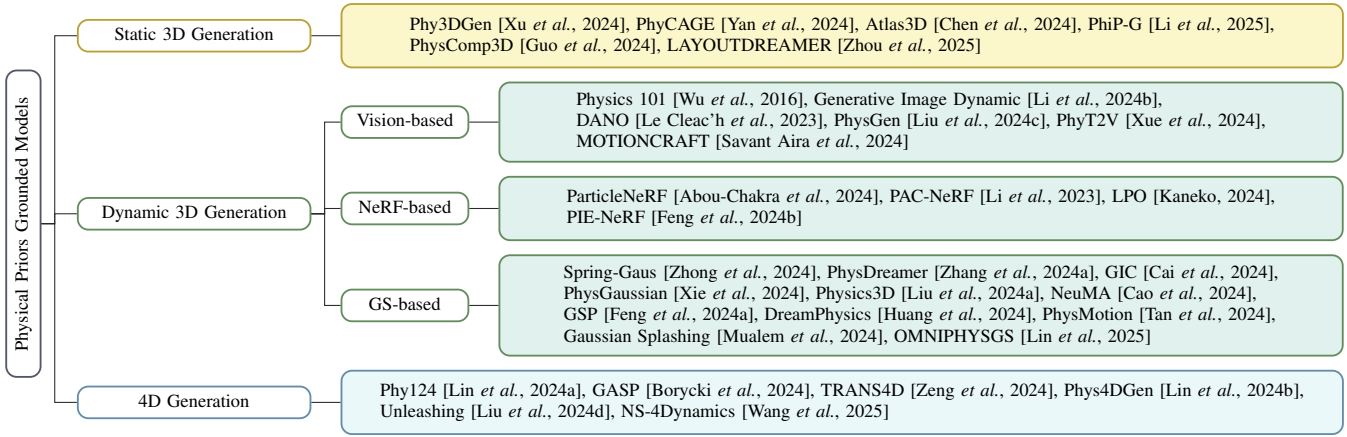


\subsubsection{Scope of This Survey}
In this survey, we first explain the foundation of generative models for 3D content (Section 2.1), and introduce common 3D representations (Section 2.2) and 4D representations methods (Section 2.3). We also provide the introduction of main physical simulation techniques, including MPM, FEM, and DiffTaichi (Section 2.4). Then, we present a taxonomy of recent research that grounding physics in AIGC, categorizing it into three key areas: static 3D generation (Section 3.1), dynamic 3D generation (Section 3.2), and 4D generation (Section 3.3). Next, we present datasets (Section 4.1) and evaluation metrics (Section 4.2) in this field, and provide a detailed comparison results of several approaches on synthetic PAC-NeRF dataset in~\autoref{table:methodComparison}. Finally, our survey discuss non-negligible challenges and possible future directions for this exciting new area of research (Section 5).

\section{Background}

\subsection{Generative Models for 3D Content Generation}

Generative models have significantly advanced 3D content creation by using deep learning to synthesize realistic structures. 
Generative Adversarial Networks (GANs) \cite{goodfellow2014generative_NeurIPS2014} and Diffusion Models \cite{ho2020denoising_neurips2020} are two key approaches, each with distinct strengths. 
Originally designed for 2D generation, GANs have been adapted to 3D by incorporating voxels, point clouds, and meshes. 
 The generator synthesizes 3D data from latent codes, while the discriminator distinguishes real from generated samples. 
Though capable of producing high-quality shapes, GANs suffer from mode collapse and training instability, issues mitigated by subsequent advances.
Meanwhile, DMs provide a more stable alternative with superior sample diversity. 
They follow a two-step denoising process: first adding Gaussian noise in forward diffusion, then iteratively removing it in reverse. 
In 3D generation, these models employ diverse representations to produce high-resolution, geometrically consistent outputs \cite{poole2022dreamfusion_iclr2023,xu2024precise-physics_arxiv2024,liu2024physics3d_arxiv2024}.

\subsection{3D Representations}
3D representations include explicit and implicit methods for modeling and rendering complex objects and scenes.
\paragraph{Explicit Methods.} Explicit methods including point clouds, meshes, and 3D Gaussian Splatting, offer diverse approaches to modeling and rendering complex objects and scenes. 
Point clouds represent objects as collections of discrete points with attributes like color and surface normals, with advanced methods like SynSin \cite{wiles2020synsin_cvpr2020} and Neural Point-based Rendering \cite{dai2020neural_cvpr2020} leveraging differentiable pipelines for optimization. 
Meshes define objects through vertices, edges, and faces in polygon networks, enabling accurate shape description and property refinement through differentiable rendering. 
3D Gaussian Splatting \cite{kerbl20233d_tog2023} employs learnable 3D Gaussian kernels optimized via multi-view supervision, providing efficient real-time rendering capabilities. 
These methods, each with their unique attributes and techniques, serve different applications while contributing to the advancement of 3D modeling and rendering technology.

\paragraph{Implicit Methods.} Implicit methods such as Signed Distance Fields (SDF) and Neural Radiance Fields (NeRF), define the shapes and boundaries of objects not through explicit geometric components, but via functions that describe spatial occupancy, enabling continuous and detailed descriptions of geometries for realistic visualizations and complex operations.
Both techniques offer unique advantages: SDFs~\cite{shim2023diffusion_cvpr2023} enable efficient rendering and precise geometric manipulations like blending and smoothing of surfaces, making them valuable for computer-aided design systems and dynamic simulations, whereas NeRFs~\cite{mildenhall2021nerf_eccv2020} enable highly photorealistic rendering of 3D scenes from novel viewpoints by modeling scenes as continuous volumetric fields within a neural network. 
These complementary approaches provide developers and researchers with powerful tools for advancing the state of 3D rendering and geometric computation.

\subsection{4D Representations}
4D representations incorporate spatiotemporal information for dynamic scene synthesis and reconstruction. 
Three notable approaches have emerged in this field: K-Planes, D-NeRF, and 4D Gaussian Splatting (4DGS).
K-Planes method represents 4D scenes by factorizing the 4D space into six planes, three for spatial dimensions and three for spatiotemporal variations \cite{fridovich2023k_cvpr2023}. K-Plances encodes 4D information by separating static and dynamic components and achieves fast optimization without relying on MLP-based decoders.
D-NeRF \cite{pumarola2021d_cvpr2021} extends the NeRF framework by conditioning the radiance field on time as an additional dimension, modeling scene dynamics through time-dependent radiance fields and deformation fields to handle temporal changes. 
Splatting methods \cite{wu4DGS_cvpr2024,yang2023real_iclr2024,zhang2024mega_arxiv2024} propose two primary approaches: deformation-based transformation of 3D Gaussian kernels in image space, and direct integration of the time dimension into 3D Gaussian kernels for temporal-spatial coherence. 
These 4DGS methods offer lightweight representations of spatiotemporal information while maintaining high-quality dynamic scene reconstructions through the compactness of Gaussian kernels.


\subsection{Physical Simulation}


While 3D and 4D representations focus on modeling static and dynamic scenes respectively, physical simulation methods are essential for understanding and predicting how these representations evolve under physical laws and material properties. 
The Material Point Method (MPM), Finite Element Method (FEM), and DiffTaichi framework employ different approaches to model and compute such dynamic material behaviors across various applications.

\paragraph{Material Point Method.} MPM combines Lagrangian material points with Eulerian Cartesian grids, where material properties like density and velocity are initially stored in particles and then interpolated onto a stationary grid~\cite{jiang2016material_siggraph2016}. 
In MPM, material is viewed as collections of particles with properties such as density, velocity, and force as continuous functions of position. These particles encode local material information and move through a stationary Eulerian grid, which facilitates the computation of spatial derivatives and enforces physical laws. 
MPM simulation computes the dynamics of continuum materials through a three-stage process: particle-to-grid transfer, grid computation, and grid-to-particle transfer. Initially, material properties are stored in Lagrangian particles and interpolated to a fixed Eulerian grid using shape functions. Then, the governing physical equations are solved on the grid including conservation of momentum and mass. After solving, the updated physical quantities are interpolated back from the grid to the particles, making the particles move and update their states.
This hybrid approach enables the computation of spatial derivatives and enforcement of physical laws on the grid, after which the updated quantities are transferred back to the particles for state updates, creating an effective cycle for simulating complex material behaviors with high precision and stability.

\paragraph{Finite Element Method.} FEM denotes a classical numerical approach which is widely used in engineering and computer graphics, tackles the simulation of deformable objects by subdividing larger systems into finite elements through space discretization~\cite{zienkiewicz2000finite_2000}. 
FEM operates by formulating boundary value problems and approximating solutions through the minimization of associated error functions using variational calculus.

\paragraph{DiffTaichi.} DiffTaichi provides comprehensive differentiable programming capabilities~\cite{hudifftaichi_iclr2020}. 
It provides a two-scale automatic differentiation system that supports megakernels, imperative programming, and flexible indexing, and simplifies the implementation and optimization of simulayion methods, particularly MPM. 
This simulator demonstrates high productivity and efficient convergence in gradient-based learning and optimization tasks.

\section{Categories of Physics Prior}
To systematically study physics-grounded AIGC methods, we introduce a taxonomy based on the spatiotemporal modeling granularity and task semantics of the generated content. 
This categorization reflects how different approaches encode and enforce physical principles across varying spatial scales and temporal resolutions, which are fundamental to the generation of physically plausible outcomes.
Building on this taxonomy, as illustrated in~\autoref{fig:taxonomy}, we divide existing methods into three categories: static 3D generation, dynamic 3D generation, and 4D generation.
Static 3D generation focuses on spatial physical plausibility under fixed configurations, such as maintaining geometric integrity or modeling the mechanical properties of objects in equilibrium.
Dynamic 3D generation extends this by introducing short-term temporal dynamics, modeling how objects evolve, move, or deform over time, often using explicit or learnable motion representations.
4D generation aims at continuous spatiotemporal synthesis, capturing long-range temporal dependencies and enforcing global physical consistency throughout dynamic evolution.
This taxonomy not only aligns with the increasing complexity of physical modeling across tasks but also offers a principled lens through which to compare methods that operate at different levels of physical abstraction.
In the following subsections, we discuss each of these three categories in details.
\subsection{Static 3D Generation with Physics }
Advancements in 3D generation have enhanced both geometric accuracy and physical plausibility. 
Early works \cite{tang2023dreamgaussian_iclr2024,yi2024gaussiandreamer_cvpr2024} focused on high-fidelity shapes, while recent approaches integrate physical realism. 
\sw{
Phy3DGen \cite{xu2024precise-physics_arxiv2024} incorporates solid mechanics into a 3D diffusion model, using a differentiable physics network for refinement. 
During the optimization phase, Phy3DGen employs sample points for pretraining and implement a co-training strategy to simultaneously optimize the geometry and physics networks.
By leveraging FEM \cite{zienkiewicz2000finite_2000} and a co-optimization strategy, it ensures both visual accuracy and physical consistency.
}

Generating physically coherent multi-component structures from a single image remains challenging. 
PhysComp3D \cite{guo2024physically_arxiv2024} optimizes physical compatibility by decomposing mechanical properties, external forces, and rest-shape geometry. 
Using plastic deformation parametrization and implicit differentiation, it refines object structures but relies on predefined material properties and mesh representations. 
Extending this, PhyCAGE \cite{yan2024phycage_arxiv2024} employs Gaussian Splatting and physics-based SDS optimization to enhance efficiency. 
By generating multi-view images and refining Gaussian positions through simulation-enhanced SDS gradients, it improves the physical coherence of 3D assets.

Text-based 3D generation faces challenges in maintaining stability due to missing physical constraints. 
Atlas3D \cite{chen2024atlas3d_arxiv2024} addresses this by integrating differentiable physics simulation into an SDS-based framework, predicting rigid body dynamics and enforcing standability and stable equilibrium losses for structural consistency. 
While effective for individual objects, it lacks scene-level physical interactions. 
To address this, PhiP-G \cite{Li2025PhiP-G_arxiv2025} introduces a physics-grounded text-to-3D scene generation framework, integrating a multi-agent text processor and a relational database-based physics pool for object interactions. 
Additionally, its physical magnet module applies vector approximations to align assets in semantically and physically consistent 3D scenes. 
\textcolor{black}{Continuing this progress, LAYOUTDREAMER \cite{zhou2025layoutdreamer_arxiv2025} leverages 3DGS to create compositional scenes by converting text into a direct scene graph, which guides objects layout and physical interactions. LAYOUTDREAMER excels in generating complex multi-object scene, allowing users to conveniently edit and expand disentangled scenes.}
\subsection{Dynamic 3D Generation}
Dynamic 3D generation focuses on capturing 3D content motion over time. According to 3D representation categories, this subsection discuss three dominant paradigms: vision-based dynamic 3D generation, NeRF-based dynamic 3D generation, and GS-based dynamic 3D generation.
\sw{
Vision-based dynamic methods typically rely on information from video sequence or single image and infer the dynamic changes of 3D structure by analyzing the movements between consecutive frames.
NeRF-based dynamic 3D generation expends NeRF representation and introduces time variables to represent motion deformation.
GS-based generation methods represents the dynamic changes of 3D scenes through Gaussians which is suitable for handing the movements of both rigid and non-rigid objects.
}
\subsubsection{Vision-based Dynamic 3D Generation}
Several works explore motion modeling through frequency analysis and optical flow. 
\sw{
Li et al. \cite{li2024generative_cvpr2024} introduce a frequency-domain motion model, integrating spatial and frequency attention within LDM UNet for motion prediction. 
Their feature pyramid softmax-splatting strategy generates, allowing the model to generate future frames based on predicted spectral volumes. 
However, this method only keeps 16 low-frequency components to reduce computational complexity, which causes distortion when showing thin objects or large amounts of content.
}
MOTIONCRAFT \cite{savant2024motioncraft_arxiv2024} extends this by mapping optical flow fields between RGB and latent space in Stable Diffusion, using physics simulators to animate input images. 
This approach improves zero-shot video generation with realistic fluid dynamics, rigid motion, and multi-agent interactions, avoiding extensive data and computation costs.

To ensure physically consistent motion, differentiable physics-based approaches explicitly model object dynamics. 
\sw{
DANO \cite{le2023differentiable_RAL2023} estimates mass, center of mass, and inertia matrix of the object from a density field, and then introduces the differentiable contact model to calculate the friction force generated by the motion collision of the object.
Their proposed Monte Carlo method computes the contact force by sampling the neural density field and computing the gradient to approximate the outward normal. 
This approach can interact with existing simulation engines to optimize the control trajectory of neural objects. 
However, DANO only considers rigid body objects, while the simulation of articulated and soft body contacts remains a research challenge.
}
Physics 101 \cite{wu2016physics101_bmvc2016} instead infers physical properties from video, categorizing them into intrinsic (unobservable) and descriptive (visually detectable) attributes. 
By combining visual recognition, physics interpretation, and world simulation, it extracts properties from unlabeled video scenarios, such as sliding, falling, and floating.

Beyond explicit physics modeling, PhysGen \cite{liu2024physgen_arxiv2024} integrates model-based physics simulation into image-to-video generation. 
It first employs GPT-4V \cite{OpenAI_GPT-4V_2023} for scene perception, extracting materials, object composition, and physical attributes from single input image. 
Rigid-body physics then simulates object dynamics, followed by motion-guided rendering for realistic and controllable training-free video synthesis.
\textcolor{black}{However, models still struggle to understand the interaction and movement of multiple objects. 
To enforce physics-grounded video generation, PhyT2V \cite{xue2024phyt2v_cvpr2025} embeds physical rules into the text prompts by local chain-of-thought (CoT) and global step-back reasoning. 
PhyT2V expends the exsiting video generation models to out-of-distribution domains with sufficient and appropriate contexts via LLM, enabling significant quality improvement.}

\subsubsection{NeRF-based Dynamic 3D Generation}
Dynamic NeRF methods \cite{fang2022fast_siggraph2022,gao2021dynamic_ICCV2021,park2021nerfies_iccv2021} require full image sequences for training, limiting adaptability to dynamic scenes. 
\sw{
ParticleNeRF \cite{abou2024particlenerf_wacv2024} addresses this by introducing a particle-based encoding, updating particle positions via backpropagated photometric loss. 
They use a lightweight physics system to manage particle collisions and motion, enabling continuous adaptation to objects deformation.
Unlike traditional static grid encodings, ParticleNeRF allows faster adaptation to dynamic scenes involving rigid bodies, articulated objects, and deformable entities with higher fidelity and efficiency.
}
Further, PAC-NeRF \cite{lipac-nerf_iclr2023} extends NeRF’s capabilities by estimating geometry and physical parameters from multi-view videos using a hybrid Eulerian-Lagrangian representation. 
It combines NeRF density fields with MPM physical simulation, enabling differentiable rendering and simulation without predefined object structures.

Despite its advantages, PAC-NeRF relies on first-frame grid representations, restricting optimization across sequences. 
\sw{
LPO \cite{kaneko2024improving_cvpr2024} overcomes this by introducing Lagrangian particle optimization, which refines particle positions and features across entire videos. 
Additionally, it incorporates physical constraints from the MPM to iteratively correct geometric structures from video sequences, which addresses the bottleneck of geometry learning in sparse-view settings.
}
While these methods enhance motion modeling, NeRF remains computationally demanding for elastic dynamics simulation, especially for complex deformable objects. 
PIE-NeRF \cite{feng2024pie-nerf_cvpr2024} addresses this by integrating physics-based simulation into NeRF, employing quadratic generalized least squares (Q-GMLS) \cite{martin2010unified_tog2010} for nonlinear dynamics and large deformations. 
Using spatial reduction and real-time neural graphics primitives (NGP) \cite{muller2022instant_tog2022}, it enables interactive manipulation and real-time rendering, achieving both physical accuracy and visual fidelity.

\subsubsection{GS-based Dynamic 3D Generation}

Traditional NeRF-based dynamic models assume known material properties, limiting abilities to simulate heterogeneous objects. 
Spring-Gaus \cite{zhong2024reconstruction_arxiv2024} introduces a 3D Spring-Mass model with learnable mass points and springs, enabling elastic reconstruction from multi-view videos. 
By integrating Gaussian kernels with the Spring-Mass model, it decouples appearance and geometry, efficiently capturing geometry, motion, and physical properties.

\sw{
To model action-conditioned dynamics, PhysDreamer \cite{zhang2025physdreamer_eccv2024} learns dynamics priors from video generation models. 
The framework distills dynamics priors through differentiable Material Point Method (MPM) simulation and rendering, optimizing physical parameters such as the Young’s modulus field, which controls object stiffness.
Additionally, to enhance computational efficiency, PhysDreamer introduces an accelerated simulation strategy by employing K-means clustering, creating “driving particles” that reduce computational overhead while maintaining physical fidelity.
}
Extending this, Physics3D \cite{liu2024physics3d_arxiv2024} integrates viscoelastic MPM with Score Distillation Sampling (SDS) to simulate a wide range of materials, from elastic to plastic behaviors. 
PhysMotion \cite{tan2024physmotion_arxiv2024} introduces physics-based simulation MPM to guide intermediate 3D representations from a single image. 
Unlike PhysDreamer and Physics3D, which focus on learning physical properties from video diffusion models, PhysMotion combines 3DGS and refines the coarse simulation using a 2D image DM with cross-frame attention.

For fluid-solid coupling, Gaussian Splashing (GSP) \cite{feng2024splashing_arxiv2024} utilizes 3D Gaussian kernels as particles, tracking fluid surfaces and interpolating deformations onto Gaussian kernels. 
By incorporating surface tension and specular effects, it achieves physically consistent rendering of solid-fluid interactions. 
In underwater dynamic modeling, Gaussian Splashing for underwater imagery \cite{mualem2024gaussian_arxiv2024} extends 3D Gaussian Splatting with learnable backscatter and attenuation effects, introducing a depth-aware rasterization pipeline for robust underwater reconstruction. Additionally, Gaussian Splashing also provides a novel underwater dataset TableDB, consisting 172 images with a resolution of 1384 $\times$ 918, which are unbounded in camera-to-scene distances.

To address discontinuities in video diffusion-based dynamics, DreamPhysics \cite{huang2024dreamphysics_arxiv2024} introduces Motion Distillation Sampling (MDS), improving motion realism over standard SDS. 
It employs KAN-based material fields and frame boosting, enabling text- and image-conditioned physical simulations without requiring ground truth videos. 
PhysGaussian \cite{xie2024physgaussian_cvpr2024} further bridges Newtonian dynamics and 3D Gaussian kernels, leveraging continuum mechanics-driven deformation models to align physical simulation and visual rendering, handling a variety of materials including metals and non-Newtonian fluids.
\textcolor{black}{However, PhysGaussian still require manual tuning physical properties, which is time-consuming and highly relies on expert knowledge.
To enhance the flexibility and generalizability of the constitutive model, OMNIPHYSGS \cite{lin2025omniphysgs_iclr2025} introduces a general physics-based 3D dynamic synthesis framework by treating each 3D asset as a collection of learnable constitutive 3D Gaussians.
Then, OMNIPHYSGS leverages a pre-trained video diffusion model~\cite{poole2022dreamfusion_iclr2023} to supervise the estimation of material weighting factors, enabling the synthesis of physically plausible dynamics across a broad spectrum of materials.}

For generalizable modeling, NeuMA \cite{cao2024neural_arxiv2024} integrates physical laws with learned corrections, introducing Neural Constitutive Laws (NCLaw) for adaptive physics simulation. 
It incorporates Particle-GS, which binds simulation particles to Gaussian kernels, improving visual-grounded dynamic modeling. 
GIC (Gaussian-Informed Continuum) \cite{cai2024gic_NeurIPS2024} further refines physical property estimation, using motion decomposition networks and a coarse-to-fine density field generation strategy, enhancing dynamic scene reconstruction and continuum mechanics-based physical optimization.

\subsection{4D Generation}
4D generation aims to reconstruct 3D presentations from input conditions such as text, image, and video sequences. Currently, most 4D generation works heavily rely on powerful 3D generation works, which have high computational costs and face challenges in understanding real-world dynamics. 

\sw{
To improve 4D generation models modeling and understanding ability, Phy124 \cite{lin2024phy124_arxiv2024} eliminates diffusion models, enabling fast, physics-driven 4D content generation by converting an image into a 3D Gaussian representation and applying MPM to simulate Gaussian field dynamics. 
First, it transforms the input image into a static 3D Gaussian representation. 
In the second stage, the Material Point Method (MPM) \cite{jiang2016material_siggraph2016,hu2018moving_tog2018} is used to simulate the physical dynamics of the 3D Gaussian field, where each Gaussian kernel is treated as a discrete material point with physical properties such as mass, density, and volume. 
By using MPM for physics-grounded dynamics, Phy124 eliminates the need for diffusion models in 4D generation, significantly speeding up the process. 
}

\sw{
Similarly to Phy124, which focuses on 4D content generation by modeling dynamic 3D objects evolving over time, GASP \cite{borycki2024gasp_arxiv2024} extends Gaussian-based modeling for real-time 3D simulation.
They leveraged the GaMeS framework \cite{waczynska2024games_arxiv2024}, utilizing flat Gaussian representations to map Gaussian components into triangle face representations, treating each 3D point as a discrete entity. 
Their method supports real-time simulations, efficiently handling both static and dynamic 3D scenes.}
Phys4DGen \cite{lin2024phys4dgen_arxiv2024} further extends Phy124 by integrating a PPM for material-aware physics simulation, segmenting material groups from images and inferring properties via GPT-4o \cite{OpenAI_GPT-4V_2023}, enhancing recognition and simulation fidelity.

For text-to-4D synthesis, TRANS4D \cite{zeng2024trans4d_arxiv2024} introduces multi-modal large language model priors to generate detailed and physically plausible 4D scene data from original textual input. 
Based on the planning 4D scene data, they calculate the 3DGS transformation function at each timestep. 
After obtaining the physics-aware planning, they use a geometry-aware Transition Network to process them to produce the final output which serves as a reference for 4D transition.
TRANS4D enables MLLMs to initialize realistic 4D scenes with multiple interacting objects and generate geometric-awareness transitions, helping to generate more realistic 4D game scenes.
Similarly, Liu et al. \cite{liu2024unleashing_arxiv2024} propose a comprehensive 4D simulation framework, integrating GPT-4 material inference with optical flow-based loss for optimizing physical properties.
By combining multi-modal foundation models and video diffusion models, they achieve high-fidelity dynamic simulation across diverse material types.

Beyond generation, understanding and reasoning about 4D dynamic scenes remain critical for enabling physics-aware AIGC.
\textcolor{black}{NS-4Dynamics \cite{Wang2025Compositional_iclr2025} serves as the first neural-symbolic models for explicit 4D scene reconstruction. 
NS-4Dynamics incorporates physical priors into the scene parsing process, further enhancing the realism and accuracy of 4D dynamic generation.
Additionally, this work proposes the SuperCLEVR-Physics dataset, designed for video question answering tasks focused on object dynamics and interaction properties, further bridging the gap between 4D generative models and physics-aware reasoning.}

\section{Benchmarks}
In this section, we discuss the benchmarks, including datasets and evaluation metrics used to evaluate physics-aware generation models. Additionally, we present a quantitative comparison of the state-of-the-art physics-aware dynamic 3D generation models on synthetic dataset.
\subsection{Datasets}
Physics-aware generative models are typically evaluated using two types of datasets: synthetic and real-world. Each type serves a distinct role in assessing model robustness and generalization.
Synthetic datasets provide controlled dynamic environments with access to ground-truth physical properties such as object mass, density, and material parameters. 
These datasets are particularly well-suited for benchmarking models in complex scenarios involving multi-object collisions, elastic or plastic deformations, and non-Newtonian fluid behavior. 
In contrast, real-world datasets capture the diversity and unpredictability of natural scenes, offering a means to evaluate a model’s ability to transfer from simulation to reality and generalize to unconstrained, noisy conditions.
\subsubsection{Synthetic Dataset}
Synthetic PAC-NeRF consists of 9 instances with deformable objects, plastics, granular, metal, and Newtonian/Non-Newtonian fluids \cite{lipac-nerf_iclr2023}. 
Each scene depicts the process of objects falling freely, colliding, and bouncing back, captured from 11 viewpoints with ground truth data generated by the MLS-MPM framework \cite{hu2018moving_tog2018}.
Synthetic Spring-Gaus includes fourteen 3D models \cite{zhong2024reconstruction_arxiv2024}, all of which are generated from PAC-NeRF \cite{lipac-nerf_iclr2023} and OmniObject3D \cite{wu2023omniobject3d_cvpr2023} approaches. 
This dataset features elastic object sequences captured from 10 viewpoints across 30 frames at a resolution of $512\times 512$.

\begin{table*}
    \setlength{\tabcolsep}{24pt} 
    \renewcommand{\arraystretch}{1.15} 
    \centering
    \scalebox{0.86}{
    \begin{tabular}{c|c}
    \toprule
         \textbf{Material Types}& \textbf{Physical Properties}  \\
    \midrule
         Newtonian fluid      & Fluid viscosity $\mu$, Bulk modulus $\kappa$ \\
         Non-Newtonian fluid  & Shear modulus $\mu$, Bulk modulus $\kappa$, Yield stress $\tau_Y$, Plastic viscosity $\eta$ \\
         Elasticity           & Young's modulus $E$, Poisson's ratio $\nu$ \\
         Plasticine           & Young's modulus $E$, Poisson's ratio $\nu$, Yield stress $\tau_Y$  \\
         Metal                & Young's modulus $E$, Poisson's ratio $\nu$, Yield stress $\tau_Y$  \\
         Foam                 & Young's modulus $E$, Poisson's ratio $\nu$, Plastic viscosity $\eta$  \\
         Sand                 & Friction angle $\theta_{fric}$  \\
    \bottomrule
    \end{tabular}}
    \caption{A taxonomy of seven common material types simulated with various physical properties.}
    \label{tab:physical_properties}
\end{table*}

\subsubsection{Real-world Dataset}
Real-world PAC-NeRF captures a deformation ball falling onto a table using a capture system comprising four synchronized Intel RealSense D455 cameras. 
The real-world data in PAC-NeRF are RGB images at a resolution of $640\times 480$ and at a rate of 60 frames per second. 
Real-world Spring-Gaus contains both static scenes and dynamic multi-view videos \cite{zhong2024reconstruction_arxiv2024}. 
Static scenes include 50-70 images from various viewpoints, while dynamic scenes are recorded from three viewpoints at a resolution of $1980\times 1080$. 
Physics 101 contains $17,408$ videos of 101 objects made of 15 different materials \cite{wu2016physics101_bmvc2016}. 
Each material category has 4 to 12 objects of different sizes and colors, with recorded physical properties such as mass, volume, and density. 
VIDEOPHY evaluates whether generated videos adhere to physical commonsense \cite{bansal2024videophy_arxiv2024}. 
It comprises 688 human-verified high-quality captions, with 344 prompts for the test set and 344 prompts for train set. 
This dataset maintains a balanced distribution of the state of matter and complexity across both sets.

\subsubsection{Physical Properties}
From the definition in MPM simulator~\cite{jiang2016material_siggraph2016}, seven material types include Newtonian and non-Newtonian fluids, elasticity, plasticine, metal, foam, and sand are characterized by specific physical parameters (\autoref{tab:physical_properties}). For instance, Newtonian fluids are defined by viscosity and bulk modulus, while Non-Newtonian fluids also consider shear modulus and yield stress. For the synthetic dataset, each object’s material type is predefined, while the real-world dataset usually lack the information of object material.

\begin{table*}[htbp]
\setlength{\tabcolsep}{12pt}
\renewcommand{\arraystretch}{1.15}
\centering
\scalebox{0.8}{
\begin{tabular}{cccccccc}
\toprule
\multirow{2}{*}{\textbf{Materials}} & \multirow{2}{*}{\textbf{Objects}} & \multirow{2}{*}{\textbf{GT}} & \multirow{2}{*}{\textbf{Parameters}} & \multicolumn{2}{c}{\textbf{NeRF-based Methods}} & \multicolumn{2}{c}{\textbf{GS-based Methods}} \\
\cmidrule(lr){5-6} \cmidrule(lr){7-8} 
& & & & \textbf{PAC-NeRF} & \textbf{LPO}  & \textbf{GIC} & \textbf{Unleashing} \\
\midrule
\multirow{3.5}{*}{Newtonian fluid} & Droplet & \begin{tabular}[c]{@{}c@{}}$200$\\ $1.0\times10^5$\end{tabular} & \begin{tabular}[c]{@{}c@{}}$\Delta\mu$\\ $\Delta\kappa$\end{tabular} & \begin{tabular}[c]{@{}c@{}}$9$\\ $8.0\times10^3$\end{tabular} & \begin{tabular}[c]{@{}c@{}}$41$\\ $2.8\times10^4$\end{tabular} & \begin{tabular}[c]{@{}c@{}}$1$\\ $8.2\times10^4$\end{tabular} & \begin{tabular}[c]{@{}c@{}}$\pmb{0.89}$\\ $\pmb{3.0\times10^3}$\end{tabular} \\
\cmidrule{2-8}
& Letter & \begin{tabular}[c]{@{}c@{}}$100$\\ $1.0\times10^5$\end{tabular} & \begin{tabular}[c]{@{}c@{}}$\Delta\mu$\\ $\Delta\kappa$\end{tabular} & \begin{tabular}[c]{@{}c@{}}$16.15$\\ $3.5\times10^4$\end{tabular} & \begin{tabular}[c]{@{}c@{}}$\pmb{2}$\\ $1.3\times10^4$\end{tabular} & \begin{tabular}[c]{@{}c@{}}$4.95$\\ $\pmb{0}$\end{tabular} & \begin{tabular}[c]{@{}c@{}}$2.27$\\ $7.0\times10^3$\end{tabular} \\
\midrule
\multirow{5.5}{*}{\begin{tabular}[c]{@{}c@{}}Non-Newtonian \\fluid\end{tabular}} 
        & Cream     
            & \begin{tabular}[c]{@{}c@{}}$1.0\times10^4$\\ $1.0\times10^6$\\ $3.0\times10^3$\\ $10$\end{tabular}   
            & \begin{tabular}[c]{@{}c@{}}$\Delta\mu$\\ $\Delta\kappa$\\ $\Delta\tau_Y$\\ $\Delta\eta$\end{tabular} 
            & \begin{tabular}[c]{@{}c@{}}$1.11\times10^5$\\ $5.7\times10^5$\\ $160$\\ $8000$\end{tabular} 
            & \begin{tabular}[c]{@{}c@{}}$2.6\times10^3$\\ $\pmb{3.2\times10^5}$\\ $40$\\ $\pmb{0.8}$\end{tabular} 
            & \begin{tabular}[c]{@{}c@{}}$\pmb{3.0\times10^2}$\\ $4.8\times10^5$\\ $\pmb{20}$\\ $3.4$\end{tabular} 
            & \begin{tabular}[c]{@{}c@{}}$1.21\times10^4$\\ $5.3\times10^5$\\ $1730$\\ $7.5$\end{tabular} \\ 
\cmidrule{2-8} 
        & Toothpaste 
            & \begin{tabular}[c]{@{}c@{}}$5.0\times10^3$\\ $1.0\times10^5$\\ $200$\\ $10$\end{tabular}    
            & \begin{tabular}[c]{@{}c@{}}$\Delta\mu$\\ $\Delta\kappa$\\ $\Delta\tau_Y$\\ $\Delta\eta$\end{tabular} 
            & \begin{tabular}[c]{@{}c@{}}$1510$\\ $5.122\times10^4$\\ $28$\\ $0.23$\end{tabular}                    
            & \begin{tabular}[c]{@{}c@{}}$340$\\ $7.88\times10^4$\\ $38$\\ $\pmb{0.20}$\end{tabular}          
            & \begin{tabular}[c]{@{}c@{}}$810$\\ $7.6\times10^4$\\ $\pmb{26}$\\ $0.90$\end{tabular}                       
            & \begin{tabular}[c]{@{}c@{}}$\pmb{70}$\\ $\pmb{1.3\times10^4}$\\ $46$\\ $28.99$\end{tabular} \\ 
\midrule
\multirow{3}{*}{Elasticity}                                                     
        & Torus      
            & \begin{tabular}[c]{@{}c@{}}$1.0\times10^6$\\ $0.3$\end{tabular}                           
            & \begin{tabular}[c]{@{}c@{}}$\Delta E$\\ $\Delta\nu$\end{tabular}                                            
            & \begin{tabular}[c]{@{}c@{}}$4.0\times10^4$\\ $0.022$\end{tabular}                                                 
            & \begin{tabular}[c]{@{}c@{}}$9.0\times10^4$\\ $0.007$\end{tabular}                            
            & \begin{tabular}[c]{@{}c@{}}$\pmb{1.0\times10^4}$\\ $\pmb{0.005}$\end{tabular}                                        
            & \begin{tabular}[c]{@{}c@{}}$3.9\times10^4$\\ $0.297$\end{tabular} \\ 
\cmidrule{2-8} 
        & Bird       
            & \begin{tabular}[c]{@{}c@{}}$3.0\times10^5$\\ $0.3$\end{tabular}                    
            & \begin{tabular}[c]{@{}c@{}}$\Delta E$\\ $\Delta\nu$\end{tabular}                                            
            & \begin{tabular}[c]{@{}c@{}}$2.2\times10^5$\\ $0.027$\end{tabular}                                                 
            & \begin{tabular}[c]{@{}c@{}}$1.9\times10^4$\\ $0.047$\end{tabular}                           
            & \begin{tabular}[c]{@{}c@{}}$\pmb{8.0\times10^3}$\\ $\pmb{0.016}$\end{tabular}                                       
            & \begin{tabular}[c]{@{}c@{}}$1.2\times10^4$\\ $0.171$\end{tabular} \\ 
\midrule
\multirow{4.5}{*}{Plasticine}                                                     
        & Playdoh    
            & \begin{tabular}[c]{@{}c@{}}$2.0\times10^6$\\ $0.3$\\ $1.54\times10^4$\end{tabular} 
            & \begin{tabular}[c]{@{}c@{}}$\Delta E$\\ $\Delta\nu$\\ $\Delta\tau_Y$\end{tabular}                        
            & \begin{tabular}[c]{@{}c@{}}$1.84\times10^6$\\ $0.028$\\ $150$\end{tabular}                                       
            & \begin{tabular}[c]{@{}c@{}}$7.2\times10^5$\\ $0.063$\\ $7600$\end{tabular}                 
            & \begin{tabular}[c]{@{}c@{}}$4.2\times10^5$\\ $\pmb{0.022}$\\ $\pmb{20}$\end{tabular}                               
            & \begin{tabular}[c]{@{}c@{}}$\pmb{2.11\times10^5}$\\ $0.098$\\ $41$\end{tabular} \\ 
\cmidrule{2-8} 
       & Cat        
           & \begin{tabular}[c]{@{}c@{}}$1.0\times10^6$\\ $0.3$\\ $3.85\times10^3$\end{tabular}        
           & \begin{tabular}[c]{@{}c@{}}$\Delta E$\\ $\Delta\nu$\\ $\Delta\tau_Y$\end{tabular}                        
           & \begin{tabular}[c]{@{}c@{}}$8.39\times10^5$\\ $0.007$\\ $280$\end{tabular}                                       
           & \begin{tabular}[c]{@{}c@{}}$8.03\times10^5$\\ $\pmb{0.003}$\\ $650$\end{tabular}                  
           & \begin{tabular}[c]{@{}c@{}}$\pmb{2.0\times10^3}$\\ $0.004$\\ $\pmb{90}$\end{tabular}                               
           & \begin{tabular}[c]{@{}c@{}}$3.87\times10^5$\\ $0.124$\\ $850$\end{tabular} \\ 
\midrule
Sand & Trophy & $40$ & $\Delta\theta_{fric}$ & $3.9^\circ$ & $2.25^\circ$ & $2.0^\circ$ & $\pmb{0.5^\circ}$ \\
\bottomrule
\end{tabular}}
\caption{Performance comparison of dynamic 3D generation methods on the synthetic PAC-NeRF dataset.}
\label{table:methodComparison}
\end{table*}


\subsubsection{Evaluation Metrics} 
Evaluating physics-prior guided generation methods requires assessing three key aspects: semantic coherency, physical consistency, and model performance, to ensure alignment with physical principles.
For semantic coherency, we adopt the CLIP score \cite{radford2021learning_icml2021} to quantify the alignment between generated content and textual descriptions by computing cross-modal embeddings. 
Additionally, we use the Semantic Adherence (SA) metric, which assigns a binary score ($\rm SA \in {0,1}$), where $\rm SA=1$ indicates that the generated video is semantically grounded in its corresponding text caption.
For physical consistency, we report the Mean Absolute Error (MAE) between the generated physical properties and ground-truth values. 
We also introduce the Physical Commonsense (PC) metric, a binary indicator ($\rm PC \in {0,1}$) denoting whether the generated content adheres to intuitive physical laws commonly understood by humans.
Lastly, for overall model performance, we use Peak Signal-to-Noise Ratio (PSNR) to evaluate the quality of generated visual content, while Structural Similarity Index Measure (SSIM) quantifies the perceptual similarity between generated and reference images, especially in dynamic scenes.

\subsubsection{Comparison Results}
We select the state-of-the-art approaches in physics-aware dynamic 3D generation, and each utilizing different representations and architectures to integrate physical priors. Specifically, PAC-NeRF \cite{lipac-nerf_iclr2023} and LPO \cite{kaneko2024improving_cvpr2024} are NeRF-based approaches, while GIC \cite{cai2024gic_NeurIPS2024} and unleashing \cite{liu2024unleashing_arxiv2024} are GS-based approaches. 
\autoref{table:methodComparison} provides a comparative analysis of four physics-aware 3D generation methods on the Synthetic PAC-NeRF dataset, which includes nine objects composed of Newtonian fluids, Non-Newtonian fluids, elasticity-based materials, plasticine, and sand. 
Besides, the dataset provides ground-truth physical simulation data, allowing for a quantitative evaluation of these methods based on key physical parameters such as viscosity, yield stress, sheer modulus, bulk modulus, elasticity modulus, Poisson’s ratio, and friction angle.

The comparison reveals that Unleashing consistently achieves the lowest AE for most fluid and plasticine materials, including Newtonian fluids (Droplet, Letter), Non-Newtonian fluids (Toothpaste), and plasticine (Playdoh, Cat). 
LPO demonstrates superior performance in yield stress modeling for Non-Newtonian fluids, achieving the best bulk modulus ($\Delta\kappa$) results for Cream. 
GIC is the strongest performer in elasticity-based materials, obtaining the lowest AE in Torus and Bird ($\Delta E$, $\Delta \nu$). 
Unleashing is competitive in estimating sand friction angle ($\Delta \theta_{\text{fric}}$) for the Trophy object, indicating its effectiveness.

Across different methods, PAC-NeRF exhibits higher errors, particularly in yield stress ($\Delta \tau_Y$) and elasticity ($\Delta E$), probably due to its fixed first-frame optimization. 
LPO refines particle positions, improving performance in Non-Newtonian fluids but struggling with elasticity-based materials. 
GIC excels in elasticity modeling, achieving the lowest $\Delta E$ values for Torus and Bird. 
Unleashing demonstrates the most robust overall performance, excelling in Newtonian and Non-Newtonian fluids as well as plasticine, showcasing its effectiveness in capturing complex material behaviors.

Overall, the results highlight the importance of integrating physics-aware priors in generative models, as different methods exhibit distinct strengths depending on material properties. 
Unleashing demonstrates the most robust performance across various material types, particularly in modeling fluid dynamics and plasticine deformations. 
GIC excels in elasticity-based materials due to its Gaussian Splatting-based scene reconstruction, effectively capturing non-rigid deformations. 
LPO refines PAC-NeRF by optimizing particle positions in time-varying sequences, improving its accuracy in Non-Newtonian fluids, though it struggles with elasticity modeling. 
PAC-NeRF, despite pioneering a hybrid Eulerian-Lagrangian representation, suffers from larger errors due to its reliance on first-frame initialization, limiting its adaptability to dynamic materials.


\section{Challenges and Future Directions} 

Despite significant progress in physics-grounded generative models, several challenges remain that hinder their broader applicability and effectiveness. 
Current approaches often struggle with accurately modeling physical interactions, maintaining long-term dynamic consistency, and generalizing to diverse materials and real-world scenarios. 
Addressing these limitations requires advancements in dataset construction, model design, and integration with physics-aware reasoning.

One major challenge is the lack of accurate physical parameter annotations in existing datasets, which limits the scalability of data-driven generative models. 
Most datasets lack expert-annotated physical properties and fail to cover a diverse range of material behaviors and dynamic interactions. 
While recent works have explored leveraging large language models (LLMs) for physical reasoning \cite{liu2024physgen_arxiv2024,lin2024phys4dgen_arxiv2024,liu2024unleashing_arxiv2024}, these approaches struggle with capturing intricate physical dependencies. 
The effectiveness of LLMs in physical reasoning heavily depends on well-constructed prompts, and current models \cite{OpenAI_GPT-4V_2023} exhibit inconsistency in physics-related tasks. 
A promising research direction is to enhance learning and reasoning capabilities within LLMs.

Another fundamental challenge lies in embedding physical constraints into generative models, particularly in understanding the relationship between dynamic movements and physical laws. 
While properties such as mass, velocity, and friction play a crucial role in realistic motion synthesis, effectively integrating these parameters into neural representations remains an open problem. 
3D Gaussian Splatting (3DGS)-based approaches \cite{xie2024physgaussian_cvpr2024,cai2024gic_NeurIPS2024,borycki2024gasp_arxiv2024,liu2024unleashing_arxiv2024} attempt to encode physical parameters by extending Gaussian kernels with additional dimensions and incorporating physics-aware optimization in differentiable networks. 
However, for 4D dynamic generation, existing modeling techniques are still underdeveloped, and further exploration is needed to improve their ability to capture complex object deformations and temporal consistency.

Future research should focus on enhancing the integration of physics-aware priors in generative models, improving their adaptability to diverse materials and dynamic environments. 
For example, advancing Sim2Real transfer and Embodied AI will further bridge the gap between simulated and real-world interactions, enabling more physically consistent and generalizable generative models.

\section{Conclusion}
This survey reviews physics-grounded generative models for 3D and 4D content generation, categorizing methods based on their representation and generation paradigms. 
We analyze their strengths and limitations, highlighting how physical priors improve visual realism and structural consistency. Quantitative comparisons reveal gaps in physical accuracy and generalization.
Despite progress, challenges remain in modeling multi-object complex physical interactions and enhancing dataset diversity in this area. 
Future research should focus on improving physics-aware learning, integrating differentiable physics, and advancing Sim2Real transfer to bridge the gap between simulation and real-world applications.

\small
\bibliographystyle{named}
\bibliography{ijcai25}

\end{document}